%% file: paper.tex
\documentclass[]{TEAI}
\usepackage{helvet}
\usepackage{amsmath} 
\usepackage{natbib}
\usepackage{graphicx}
\usepackage{caption}
\usepackage{subcaption} 
\usepackage[toc,page,header]{appendix}
\usepackage[utf8]{inputenc} 
\usepackage[T1]{fontenc}    
\usepackage{hyperref}       
\usepackage{url}            
\usepackage{booktabs}       
\usepackage{lmodern}        
\usepackage{amsfonts}       
\usepackage{nicefrac}       
\usepackage{microtype}      
\usepackage{wrapfig}
\usepackage{amsmath}
\usepackage{bbm} 
\usepackage{amssymb}  
\usepackage{fontawesome}  
\usepackage{url}  

\usepackage{titletoc}

\usepackage{tikz}  
\usepackage{comment}  
\usepackage{tabularx}  
\usepackage{booktabs}  

\usepackage{minitoc}

\usepackage{booktabs}
\usepackage{array}
\usepackage{etoolbox}

\definecolor{lightblue}{RGB}{200, 230, 255}  
\definecolor{headerblue}{RGB}{150, 200, 255} 

\usepackage{pgfplots}
\usepackage[utf8]{inputenc} 
\usepackage[T1]{fontenc}    
\usepackage{hyperref}       
\usepackage{url}            
\usepackage{booktabs}       
\usepackage{amsfonts}       
\usepackage{nicefrac}       
\usepackage{microtype}      
\usepackage{xcolor}         
\usepackage{graphicx}
\usepackage{float}
\usepackage{comment}
\usepackage{multirow} 
\usepackage{amsmath} 
\usepackage{makecell} 
\usepackage{siunitx}  
\usepackage{tikz}
\usepackage{pgf-pie} 
\usepackage{subcaption}
\usepackage{wrapfig}
\usepackage[export]{adjustbox}

\usepackage{ragged2e}      
\usepackage{tabularx}       
\usepackage{array}          
\usepackage{caption}        
\usepackage{enumitem}
\usepackage{pifont}
\usepackage[hang,flushmargin]{footmisc} 

\usepackage{tcolorbox}

\usepackage{tcolorbox}    
\tcbuselibrary{breakable}  
\tcbuselibrary{skins}      

\usepackage{tabularx}
\usepackage{listings}

\title{Thinking Traps in Long Chain-of-Thought: A Measurable Study and Trap-Aware Adaptive Restart}

\author{
    Kang Chen\textsuperscript{1,*},
    Fan Yu\textsuperscript{1,*},
    Junjie Nian\textsuperscript{1},
    Shihan Zhao\textsuperscript{1},
    Zhuoka Feng\textsuperscript{1},
    Zijun Yao\textsuperscript{},
    Heng Wang\textsuperscript{1},
    Minshen Yu\textsuperscript{1},
    Yixin Cao\textsuperscript{1,2,$\dagger$}
}

\affiliation[1]{\mbox{Fudan University}}
\affiliation[2]{\mbox{Shanghai Innovation Institute}}

\input{section/0_abstract/abstract}

\correspondence{\email{yxcao@fudan.edu.cn}}

\begin{document}
\maketitle
\renewcommand{\thefootnote}{}
\footnotetext{$^*$Equal Contribution.\\$^\dagger$Corresponding authors.}
\renewcommand{\thefootnote}{\arabic{footnote}}


\vspace{-1.5em}

\input{section/1_introduction/introduction}
\input{section/2_related_works/related_works}
\input{section/3_problem_formulation/problem_formulation}
\input{section/4_method/method}
\input{section/5_experiment/experiment}
\input{section/6_analysis/analysis}

\input{section/7_discussion/discussion}
\input{section/8_limitations/limitation}

\clearpage
\bibliographystyle{plainnat}
\bibliography{main}
\input{section/9_appendix/appendix}
\end{document}

%% file: section/0_abstract/abstract.tex
\abstract{
\begin{abstract}

Scaling test-time compute via Long Chain-of-Thought (Long-CoT) significantly enhances reasoning capabilities, yet extended generation does not guarantee correctness: after an early wrong commitment, models may keep elaborating a self-consistent but incorrect prefix.
Through fine-grained trajectory analysis, we identify \textbf{Thinking Traps}, prefix-dominant deadlocks where later reflection, alternative attempts, or verification fails to revise the root error. On a curated subset of DAPO-MATH, 89\% of failures exhibit such traps. To solve this problem, we introduce \textbf{TAAR} (Trap-Aware Adaptive Restart), a test-time control framework that trains a diagnostic policy to predict two signals from partial trajectories: a trap index for \emph{where} to truncate and an escape probability for \emph{whether and how strongly} to intervene.
At inference time, TAAR truncates the trajectory before the predicted trap segment and adaptively restarts decoding; for severely trapped cases, it applies stronger perturbations, including higher-temperature resampling and an optional structured reboot suffix.
Experiments on challenging mathematical and scientific reasoning benchmarks (AIME24, AIME25, GPQA-Diamond, HMMT25, BRUMO25) show that TAAR improves reasoning performance without fine-tuning base model parameters.

\end{abstract}
}

%% file: section/1_introduction/introduction.tex
\section{Introduction}

Recently, large reasoning models (LRMs) such as DeepSeek-R1~\cite{guo2025deepseek} and OpenAI o1~\cite{jaech2024openai} leverage Long Chain-of-Thought (Long-CoT) to scale test-time computation. By allocating more tokens, long traces can expose latent structure, enable step-by-step verification, and improve performance on difficult problems. However, longer reasoning also demands the ability to revisit and revise early assumptions; otherwise, additional compute may reinforce incorrect reasoning rather than improve correctness. This raises a practical question: when does ``thinking longer'' genuinely help, and when does it waste computation?

\begin{wrapfigure}{r}{0.5\columnwidth}
    \vspace{-20pt} 
    
    \centering
    \includegraphics[width=\linewidth]{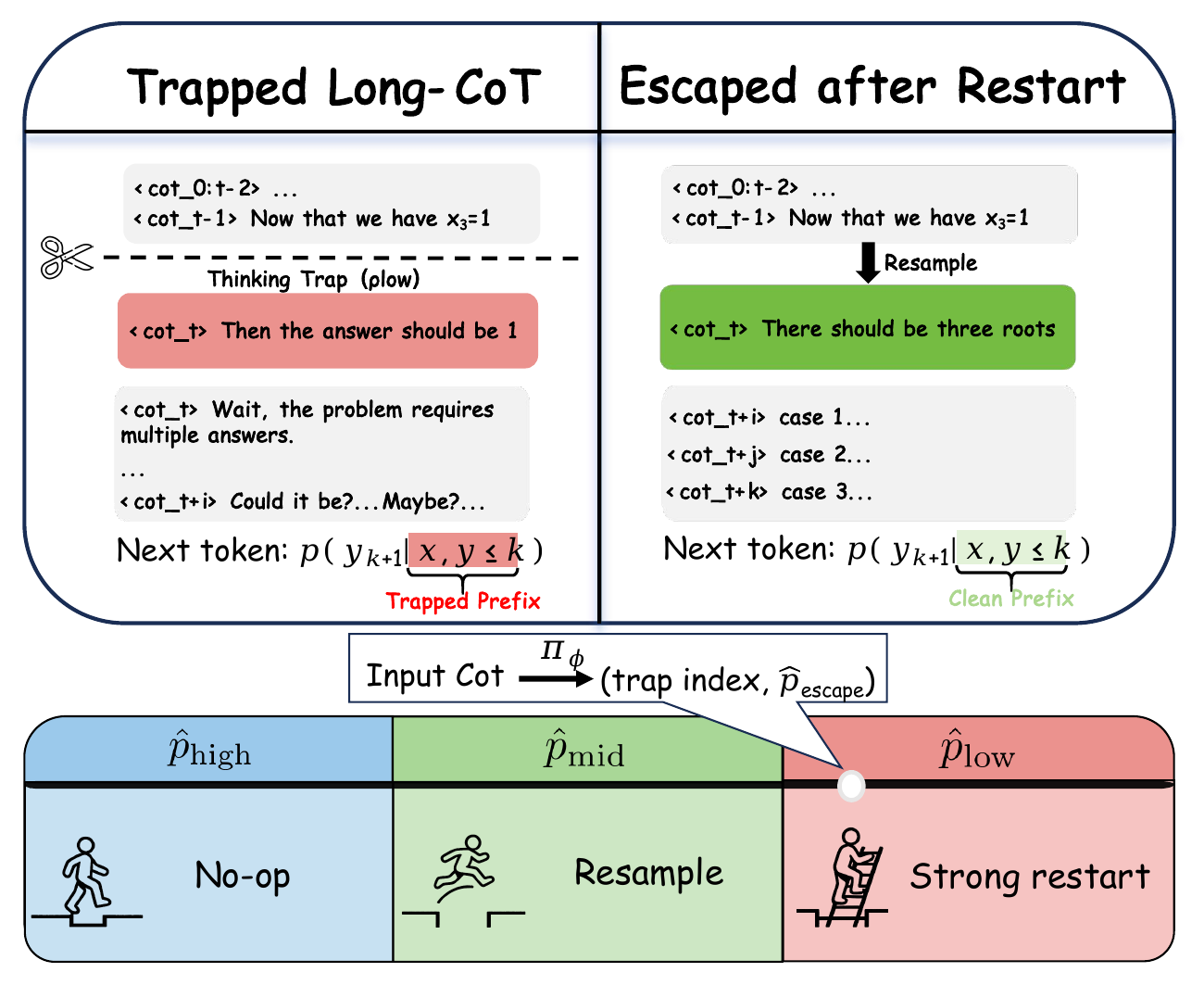}
    
    \caption{Conceptual illustration of Thinking Traps. A reasoner facing a trap can leverage Diagnostic Policy Model to choose an appropriate escape strategy: step over (no intervention), jump (mild intervention), or use a ladder (strong intervention).}
    \label{fig:concept}
    
    \vspace{-10pt} 
\end{wrapfigure}

Through fine-grained trajectory analysis of Long-CoT failures, we identify a recurring structural pattern~\cite{ding2025thinking} that explains much wasted compute. In many failed traces, an early wrong commitment dominates the continuation. Even when the model later attempts reflection or verification, these efforts often fail to revise the root cause and instead elaborate a self-consistent but incorrect prefix. We term this prefix-dominant deadlock a \textbf{Thinking Trap} (Figure~\ref{fig:concept}). Our pilot study on DAPO-MATH-17K~\cite{dapo_math_17k} finds that thinking traps are pervasive: 89\% of reasoning errors across four models exhibit such traps (Section~\ref{sec:data_construction}). Crucially, traps are not merely another error category but a bottleneck for test-time scaling, consuming substantial token budgets without yielding correctness.

To study and mitigate this phenomenon, we build a controlled framework for quantifying thinking traps. We curate a hard subset of DAPO-MATH-17K~\cite{dapo_math_17k} and collect Long-CoT trajectories from four reasoning models~\cite{qwen2025qwen3_4b_instruct_2507,deepseekai2025deepseekr10528qwen38b,openai2025gptoss20b,openai2025gptoss120b}. Each trajectory is segmented into an indexable sequence. We operationalize the trap index as the earliest segment containing a wrong commitment, identify self-repair windows where the model attempts verification, and define an escape probability via compute-matched resampling and automatic verification. This framework enables measurable study of trap prevalence, position, diagnosability, and escape behavior under fixed test-time budgets.

Building on this framework, we propose Trap-Aware Adaptive Restart (TAAR), a diagnostic-guided intervention strategy. TAAR trains a lightweight policy to predict two signals from partial trajectories: a trap index $\hat{t}$ for truncation and an escape probability $\hat{p}$ for gating intervention strength. At inference, TAAR truncates before the predicted trap segment $s_{\hat{t}}$ and adaptively restarts generation (Figure~\ref{fig:concept}), applying stronger perturbations (higher temperature or structured reboot suffix) for severe traps with low $\hat{p}$.

Experiments on five reasoning benchmarks validate TAAR improves accuracy and token efficiency without fine-tuning base models. Our main contributions can be summarized:

\begin{itemize}
\item We build a controlled, measurable study settings for thinking traps in long-CoT failures.
\item We propose TAAR, a trap-escape strategy to decide where and how strongly to intervene.
\item We have conducted systematic controlled experiments to validate the causal role of removing thinking traps and to demonstrate performance and token-efficiency gains on challenging benchmarks.
\end{itemize}

%% file: section/2_related_works/related_works.tex
\section{Related Work}

\subsection{Analysis of Long-CoT Reasoning Limitations}
\input{section/tables/table_trap_statistic}
Long-CoT scales reasoning capabilities~\cite{guo2025deepseek, jaech2024openai} but creates a vulnerability to hallucination snowballing''~\cite{zhang2023language, xu2024hallucination}. In this phenomenon, a single early deviation cascades into a coherent but factually incorrect narrative. Recent studies attribute this to "Prefix Dominance"~\cite{luo2025learning} or "Thought Anchors"~\cite{bogdan2506thought}, where initial mistakes rigidly constrain the model's attention mechanism. Consequently, models tend to rationalize their errors rather than correct them, making standard self-correction less effective~\cite{huang2023large, stechly2023gpt}. While Process Reward Models (PRMs) provide step-level verification~\cite{lightman2023let, wang2024math}, they often fail to identify errors that are locally logical but globally flawed due to a corrupted premise. We define this recursive deadlock as a Thinking Trap, positing that effective recovery requires pruning the history rather than merely continuing generation.

\subsection{Interventions during Inference}
Inference strategies have evolved from sampling-based approaches to active structural control. While massive repeated sampling~\cite{wang2022self, brown2024large} and structured search algorithms like Tree of Thoughts~\cite{yao2023tree} demonstrate that scaling test-time compute follows specific scaling laws, they often incur high costs via brute-force coverage or complex state management. Consequently, recent works propose more targeted interventions: parallel reasoning via thought exchange~\cite{luo2025learning} or inserting prompts to stimulate deeper deliberation~\cite{zhang2025smartswitch}. Other methods manipulate the reasoning medium itself: selecting optimal languages~\cite{zhang2024autocap} or injecting cross-lingual perturbations at high language uncertainty points~\cite{li2025impact}.
TAAR synthesizes these insights into a framework of diagnostic control. Distinct from untargeted or heuristic perturbations, our core innovation is trap localization: explicitly identifying the \emph{trap segment} to enable precise intervention. By truncating the corrupted effective prefix at its source and selectively triggering recovery, TAAR ensures computation is efficiently allocated to fixing root errors rather than continuing on flawed paths.

%% file: section/tables/table_trap_statistic.tex
\begin{wraptable}{r}{0.5\columnwidth}
    \vspace{-15pt} 
    \centering
    
    \resizebox{\linewidth}{!}{%
        \begin{tabular}{l|cc}
        \toprule
        \textbf{Inference Model} & \textbf{\# Errors} & \textbf{Trap Ratio (\%)} \\
        \midrule
        Qwen3-4B-Instruct & 169 & 92.90 \\
        DeepSeek-R1-Distill-Qwen-8B & 121 & 91.74 \\
        GPT-OSS-20B & 108 & 87.04 \\
        GPT-OSS-120B & 86 & 80.23 \\
        \midrule
        Total & 484 & 89.05 \\
        \midrule
        \multicolumn{3}{c}{\textit{Source: DAPO-MATH Subset (Section~\ref{sec:data_construction})}} \\
        \bottomrule
        \end{tabular}%
    }
    
    \vspace{-5pt} 
    \caption{Prevalence of Thinking Traps. We report the total number of errors (\textbf{\# Errors}) and the percentage of errors classified as Thinking Traps (\textbf{Trap Ratio}) on DAPO-sample for prevalence analysis.}
    \label{tab:trap_prevalence}
    \vspace{-10pt} 
\end{wraptable}

%% file: section/3_problem_formulation/problem_formulation.tex
\section{Thinking Traps and Escape Probability}
\label{sec:trap_formulation}

This section provides formal definitions of the core concepts that underpin TAAR. The operational procedures for obtaining these labels are described in Section~\ref{sec:data_construction}.

\subsection{Trap Index as a Wrong Commitment}
\label{sec:trap_index}
Given an input $x$, the model generates a Long-CoT trace $Y$. We structure this trace as a discrete trajectory $Y=(s_1,\ldots,s_T)$ using a segmentation function that splits the text at natural paragraph boundaries (mainly by \verb|\n\n|; details in Appendix~\ref{app:segmentation}). This segmentation transforms the continuous stream into an indexable sequence, enabling precise localization-based interventions.

We define \textbf{trap index} $t^*$ as the index of the earliest segment containing a \emph{wrong commitment}: an erroneous assumption, unjustified leap, or improper simplification that substantially restricts future reasoning. Note that $t^*$ is distinguishable from minor arithmetic slips; it acts as a structural \emph{branching point} where the trajectory becomes \emph{prefix-dominant}. Once the wrong commitment is anchored, subsequent computation tends to refine the error's consequences rather than revise the root cause. If no such error occurs, we set $t^*=\varnothing$.


\subsection{Self-Repair Windows and Escape Probability}
\label{sec:escape_probability}
Even after a wrong commitment (identified by $t^*$), the model may actively attempt to correct itself. We identify a set of segments $W \subseteq \{t^*+1,\ldots,T\}$ as \textbf{self-repair windows}. To ensure precision, we include a segment in $W$ only if it explicitly challenges the trap assumption or its consequences (e.g., through verification or by proposing an alternative approach). Routine downstream calculations that do not address the root error are excluded.

While $t^*$ localizes the error, it does not determine the severity of the deadlock. To quantify the likelihood of recovery, we define the \textbf{escape probability} $p_{\text{escape}} \in [0,1]$. Intuitively, this metric answers: \emph{if we truncate the trajectory at a self-repair attempt and resample the continuation, how often does the model succeed?}
Formally, given a verifier $\textsc{Correct}(\cdot)$ and a budget of $N$ resampled trials from valid cut points (prioritizing $W$), we estimate:
\begin{equation}
\label{eq:escape_prob}
p_{\text{escape}} \;=\; \frac{1}{N}\sum_{n=1}^{N} \mathbbm{1}\!\left[\textsc{Correct}\!\left(\hat{y}^{(n)}\right)\right],
\end{equation}
where each $\hat{y}^{(n)}$ is a continuation sampled from the truncated prefix. High $p_{\text{escape}}$ implies the trap is shallow (escapable via resampling), while low $p_{\text{escape}}$ indicates a deep deadlock requiring stronger intervention.
When $W$ is empty or provides insufficient distinct cut points, we supplement with random post-trap cut points (Section~\ref{sec:data_construction}).

%% file: section/4_method/method.tex
\section{Trap-Aware Adaptive Restart (TAAR)}
\label{sec:taar}

\begin{figure*}[t]
    \centering
    \includegraphics[width=\textwidth]{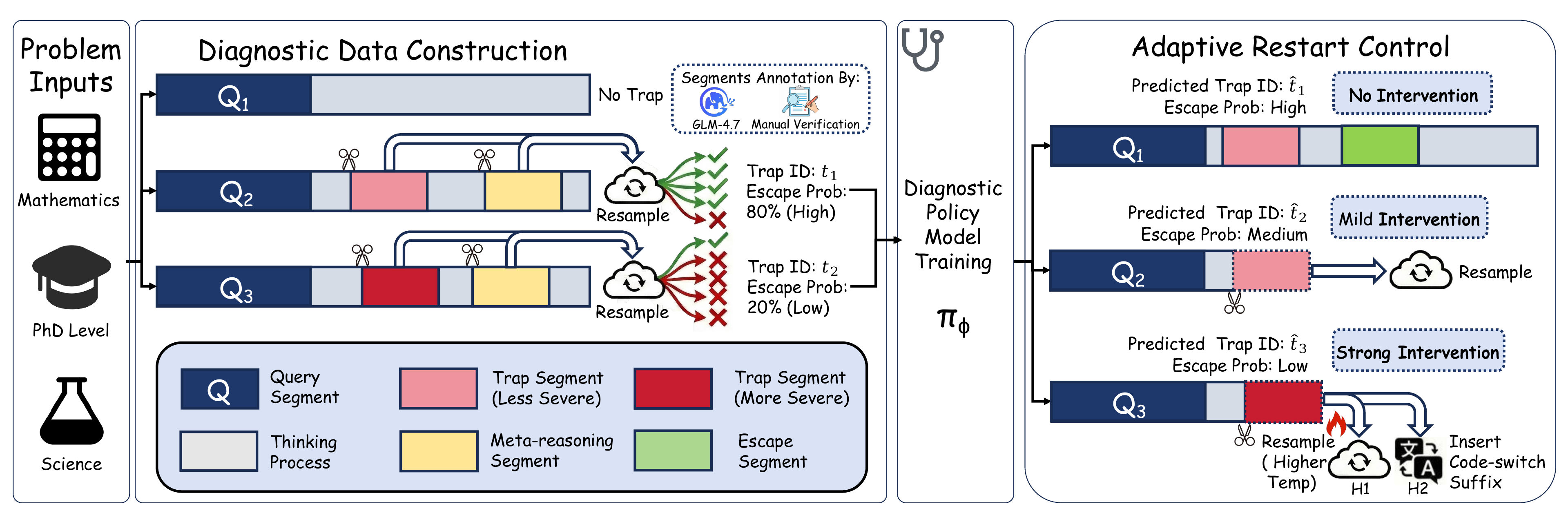}
    \caption{Overview of the TAAR framework. \textbf{Left}: Diagnostic Data Construction pipeline that segments trajectories and labels trap indices and escape probabilities via GLM-4.7 annotation with manual verification. \textbf{Middle}: Training of the diagnostic policy $\pi_\phi$. \textbf{Right}: Adaptive restart controller selects intervention based on $\hat{p}$.}
    \label{fig:framework}
\end{figure*}

We now present TAAR, a test-time control framework that operationalizes the trap diagnostics defined in Section~\ref{sec:trap_formulation}. TAAR reallocates compute away from trapped continuations toward counterfactual re-derivations (Figure~\ref{fig:framework}). The framework comprises two components: (i) a \emph{diagnostic policy} that predicts $(t^*, p_{\text{escape}})$ from partial reasoning, and (ii) an \emph{adaptive restart controller} that maps these predictions to intervention strategies.

\subsection{Control Mechanism}
\label{sec:taar_overview}

Given an instance $x$ and a (partial or complete) segmented trajectory $Y=(s_1,\ldots,s_T)$, TAAR predicts $(\hat{t},\hat{p})$, where $\hat{t}$ estimates the trap index and $\hat{p}$ estimates the escape probability. These two signals control restart decisions:

\paragraph{Where to restart.}
If intervention is triggered, TAAR truncates the trajectory \emph{before} the predicted trap segment, keeping prefix $Y_{<\hat{t}}=(s_1,\ldots,s_{\hat{t}-1})$ and regenerating a continuation from that prefix.

\paragraph{How strongly to restart.}
Not all traps are equally severe. TAAR uses $\hat{p}$ as a control signal to choose a restart operator: mild restarts encourage light exploration (e.g., default-temperature resampling), while strong restarts apply stronger perturbations (e.g., higher-temperature resampling with an optional structured reboot suffix).

\subsection{Dataset Construction}
\label{sec:data_construction}

Training the diagnostic policy requires a dataset annotated with trap indices and escape probabilities. We construct this dataset through a pipeline of trajectory generation, offline LLM annotation with human verification, and Monte Carlo estimation.

\paragraph{Source Trajectories and Annotation.}
We build a challenging subset (``DAPO-hard'') from DAPO-MATH-17K, collecting $6{,}000$ Long-CoT trajectories from four reasoning models (4B to 120B) on $1{,}500$ problems where not all models succeed. Each trajectory is segmented into an indexable sequence based on paragraph boundaries. We then employ an offline LLM judge (GLM-4.7~\cite{zai_glm47}) to analyze each segmented trace. To ensure annotation quality, two independent human annotators verify 100 random instances sampled from the LLM judgments, achieving 93\% agreement. The judge identifies the trap index $t^*$ and extracts valid self-repair windows $W$ where the model attempts verification. Ground-truth answers are provided to the judge solely to enhance offline annotation precision; TAAR does not utilize ground truth during test-time inference.

\paragraph{Escape Probability Estimation.}
To quantify the severity of the identified traps, we estimate the escape probability $p_{\text{escape}}$ via compute-matched resampling. For each trajectory, we generate $N=36$ continuations from prefixes truncated at the identified self-repair windows $W$ (supplemented by random post-trap points if $W$ is sparse). We verify these continuations using an automatic verifier (math-verify) under a fixed compute budget (temperature 0.7, max 32k tokens). This process yields a robust empirical estimate of the trajectory's ability to self-repair via plain continuation.  To ensure training quality, we further apply filtering criteria to select high-quality training samples; details are provided in Appendix~\ref{app:filtering_details}.

\subsection{Diagnostic Policy Model}
\label{sec:policy_model}

We train a policy model $\pi_\phi$ to output (i) a distribution over segment indices for trap localization, and (ii) an escape score for $\hat{p}$. The policy input concatenates the problem statement and the segmented reasoning prefix with segment labels, enabling pointer-style localization.

\paragraph{Training Setup.}
We supervise $\pi_\phi$ using the offline labels $(t^*,p_{\text{escape}})$ from \S\ref{sec:data_construction}. To make localization robust to varying amounts of post-trap ``idling'', we apply \emph{random truncation augmentation}: for each labeled trajectory, we sample an offset $\delta$ and provide the prefix up to $t^*+\delta$ as input. Formally,
{\small
\begin{equation}
x_{\text{diag}} = \mathcal{T}_{\text{in}}(x, Y_{1:t^*+\delta}),
\quad
y_{\text{diag}} = \mathcal{T}_{\text{out}}(t^*, p_{\text{escape}})
\label{eq:diag_pair}
\end{equation}
}
where $\mathcal{T}_{\text{in}}(\cdot)$ and $\mathcal{T}_{\text{out}}(\cdot)$ are formatting templates (Appendix~\ref{app:templates}).

\subsection{Adaptive Restart Controller}
\label{sec:restart_controller}

At test time, TAAR operates under a fixed compute budget (e.g., a limited number of sampled paths). Given $(\hat{t},\hat{p})$, we choose among three intervention strengths: \textbf{No Intervention:} if $\hat{p}$ is high, the trajectory is likely to self-repair; we keep the current continuation. \textbf{Mild Intervention:} if $\hat{p}$ is moderate, we restart from $Y_{<\hat{t}}$ and resample with the default decoding configuration. \textbf{Strong Intervention:} if $\hat{p}$ is low, we restart from $Y_{<\hat{t}}$ and apply a stronger perturbation, such as higher-temperature resampling and an optional \emph{structured reboot suffix}. The suffix is written in the \emph{same language as the prompt} (English in our main experiments) and explicitly requests (i) re-derivation from scratch and (ii) a checklist of key constraints before finalizing the answer. We provide the exact suffix template in Appendix~\ref{app:reboot_suffix}.

Concretely, we use $\hat{p} \geq 0.6$ for no intervention, $0.1 < \hat{p} < 0.6$ for mild intervention (resample), and $\hat{p} \leq 0.1$ for strong intervention (temperature$=$1.0 or reboot suffix). 

%% file: section/5_experiment/experiment.tex
\section{Experiments}

In this section, we evaluate the effectiveness of the \textbf{TAAR} framework. We first describe the experimental setup in \S\ref{subsec:setup}. Then, we present the main results in \S\ref{subsec:main_results}. 

\subsection{Experimental Setup}
\label{subsec:setup}
\input{section/tables/table_main}

\textbf{Base Reasoning Models.} We evaluate on the same four reasoning models used for trajectory generation (Section~\ref{sec:data_construction}): \texttt{Qwen3-4B-Instruct}~\cite{qwen2025qwen3_4b_instruct_2507}, \texttt{DeepSeek-R1-Distill-Qwen-8B}~\cite{deepseekai2025deepseekr10528qwen38b}, \texttt{GPT-OSS-20B}~\cite{openai2025gptoss20b}, and \texttt{GPT-OSS-120B}~\cite{openai2025gptoss120b}, spanning 4B to 120B parameters.

\textbf{Policy Model.} We use \texttt{Qwen3-4B-Instruct} as the backbone for the diagnostic policy $\pi_\phi$. The model is fine-tuned via supervised fine-tuning (SFT) on the dataset constructed in Section~\ref{sec:data_construction}, comprising $3{,}661$ trajectories with trap indices and escape probability labels. To enable dynamic CoT window handling, we augment the dataset via upsampling, resulting in $16{,}748$ training instances. We use LlamaFactory for training on 8$\times$H20 GPUs with learning rate $1\text{e-}5$, full fine-tuning, epoch$=$1, and maximum sequence length of 36k tokens.

\textbf{Evaluation Benchmarks.} We evaluate on five challenging reasoning benchmarks: AIME24~\cite{aime24}, AIME25~\cite{aime25}, GPQA-Diamond~\cite{rein2023gpqa}, HMMT25~\cite{matharena_hmmt_feb_2025}, and BRUMO25~\cite{matharena_brumo_2025}.

\textbf{Baselines.} We compare TAAR against the following baselines: \textbf{Base Long-CoT (Avg@4):} Standard independent sampling with 4 trajectories per problem, reporting average accuracy. \textbf{PRM:} We use \texttt{Qwen2.5-Math-PRM-7B}\footnote{\url{https://qwenlm.github.io/blog/qwen2.5-math-prm/}}~\cite{zhang2025lessons} to score 4 trajectories by averaging step-level rewards, then select the highest-scoring candidate. \textbf{AutoCap:} Adaptive routing from~\cite{zhang2024autocap} that dynamically selects the optimal reasoning language or capability based on input problem distribution.

\subsection{Main Results}
\label{subsec:main_results}

Table~\ref{tab:main_results} reports accuracy on five challenging reasoning benchmarks under a matched test-time compute budget ($N=4$ sampled trajectories). TAAR consistently improves over standard multi-sample averaging (Avg@4) for small and mid-scale models: +1.7 points on 4B model and +4.6 points on 8B model on average. Gains are most pronounced on the hardest math benchmarks (+7.5 on HMMT25, +5.9 on BRUMO25 for the 8B model), suggesting that trap-aware restart reallocates compute from trapped continuations toward genuinely different solution paths.

Compared with outcome-based selection (PRM@4), TAAR remains competitive or stronger on the mid-scale setting (+3.1 average points on 8B) without modifying base model parameters. For the 120B model, TAAR is competitive but does not consistently outperform Avg@4 or PRM@4, indicating that when base reasoning is already strong, the benefit of restart is limited by imperfect trap localization and the cost of perturbing near-correct prefixes. We further analyze these effects in Section~\ref{sec:analysis}.

%% file: section/tables/table_main.tex
\begin{table*}[t]
\centering
\small
\setlength{\tabcolsep}{4pt}
\newcommand{\gc}{\cellcolor{gray!10}}

\resizebox{\textwidth}{!}{
\begin{tabular}{l|lcccccc}
\toprule
\multirow{2}{*}{\textbf{Inference Model}} & \multirow{2}{*}{\textbf{Method}} & \textbf{AIME 24} & \textbf{AIME 25} & \textbf{BRUMO 25} & \textbf{HMMT 25} & \textbf{GPQA} & \textbf{Avg.} \\
 & & \textbf{Acc (\%)} & \textbf{Acc (\%)} & \textbf{Acc (\%)} & \textbf{Acc (\%)} & \textbf{Acc (\%)} & \textbf{(\%)} \\
\midrule
\multirow{5}{*}{\textbf{Qwen3-4B-Instruct}}
 & AVG@4 & 59.2 & \textbf{44.2} & 54.2 & 28.3 & 58.5 & 48.9 \\
 & PRM@4 & \underline{60.9} & \underline{43.5} & \textbf{57.1} & \underline{29.0} & \underline{59.7} & \underline{50.0} \\
 & Autocap & 60.0 & 36.7 & 46.7 & \textbf{33.3} & 58.1 & 47.0 \\
 & \gc \textbf{TAAR (Ours)} & \gc \textbf{64.2} & \gc \textbf{44.2} & \gc \underline{55.0} & \gc 27.5 & \gc \textbf{62.0} & \gc \textbf{50.6} \\
\midrule
\multirow{5}{*}{\textbf{DeepSeek-R1-Distill-Qwen-8B}}
 & AVG@4 & 75.0 & 67.5 & \underline{68.3} & 50.0 & 61.7 & 64.5 \\
 & PRM@4 & \underline{75.5} & \textbf{71.0} & 68.1 & \underline{51.2} & \underline{64.1} & \underline{66.0} \\
 & Autocap & 70.0 & 63.3 & 66.7 & 46.7 & 59.6 & 61.3 \\
 & \gc \textbf{TAAR (Ours)} & \gc \textbf{80.0} & \gc \underline{69.2} & \gc \textbf{74.2} & \gc \textbf{57.5} & \gc \textbf{64.5} & \gc \textbf{69.1} \\
\midrule
\multirow{5}{*}{\textbf{GPT-OSS-20B}}
 & AVG@4 & 75.8 & 75.0 & 75.8 & \underline{49.2} & \underline{59.3} & 67.0 \\
 & PRM@4 & 76.0 & \underline{77.4} & \underline{76.6} & 44.3 & 58.8 & 66.6 \\
 & Autocap & \textbf{80.0} & 76.7 & 70.0 & \textbf{53.3} & \textbf{61.1} & \underline{68.2} \\
 & \gc \textbf{TAAR (Ours)} & \gc \underline{78.3} & \gc \textbf{77.5} & \gc \textbf{80.8} & \gc 46.7 & \gc 59.0 & \gc \textbf{68.5} \\
\midrule
\multirow{5}{*}{\textbf{GPT-OSS-120B}}
 & AVG@4 & \textbf{89.2} & 84.2 & \textbf{86.7} & 68.3 & \underline{74.7} & \textbf{80.6} \\
 & PRM@4 & \underline{87.3} & 86.0 & \underline{85.5} & 69.0 & 73.8 & \underline{80.3} \\
 & Autocap & 83.3 & \underline{86.7} & 76.7 & \textbf{73.3} & \textbf{75.8} & 79.2 \\
 & \gc \textbf{TAAR (Ours)} & \gc 84.2 & \gc \textbf{87.5} & \gc 81.7 & \gc \underline{69.2} & \gc 73.4 & \gc 79.2 \\
\bottomrule
\end{tabular}
}
\caption{Main results on five challenging benchmarks. All methods operate under a fixed computational budget of $K=4$ paths. \textbf{TAAR} results are highlighted in gray. \textbf{Bold}: best; \underline{Underline}: second best.}
\label{tab:main_results}
\end{table*}

%% file: section/6_analysis/analysis.tex
\section{Analysis}
\label{sec:analysis}

\subsection{RQ1: Where and When Should We Restart?}
\label{subsec:rq1_where_restart}

The core design principle of TAAR is that effective recovery requires removing the erroneous commitment from the effective prefix, rather than attempting downstream self-correction. We define $p_{\text{escape}}$ as the proportion of restarted trajectories that reach the correct answer under a fixed resampling budget. Under a compute-matched setting, we compare three cut-point strategies: \textbf{Cut@Trap} (truncate at the predicted trap index), \textbf{Cut@Post-trap} (truncate at post-trap self-repair windows such as reflection/verification segments), and \textbf{Cut@Random} (truncate at uniformly sampled positions).

\begin{figure*}[t] 
    \centering
    \begin{minipage}[t]{0.48\textwidth}
        \centering
        \includegraphics[width=\linewidth]{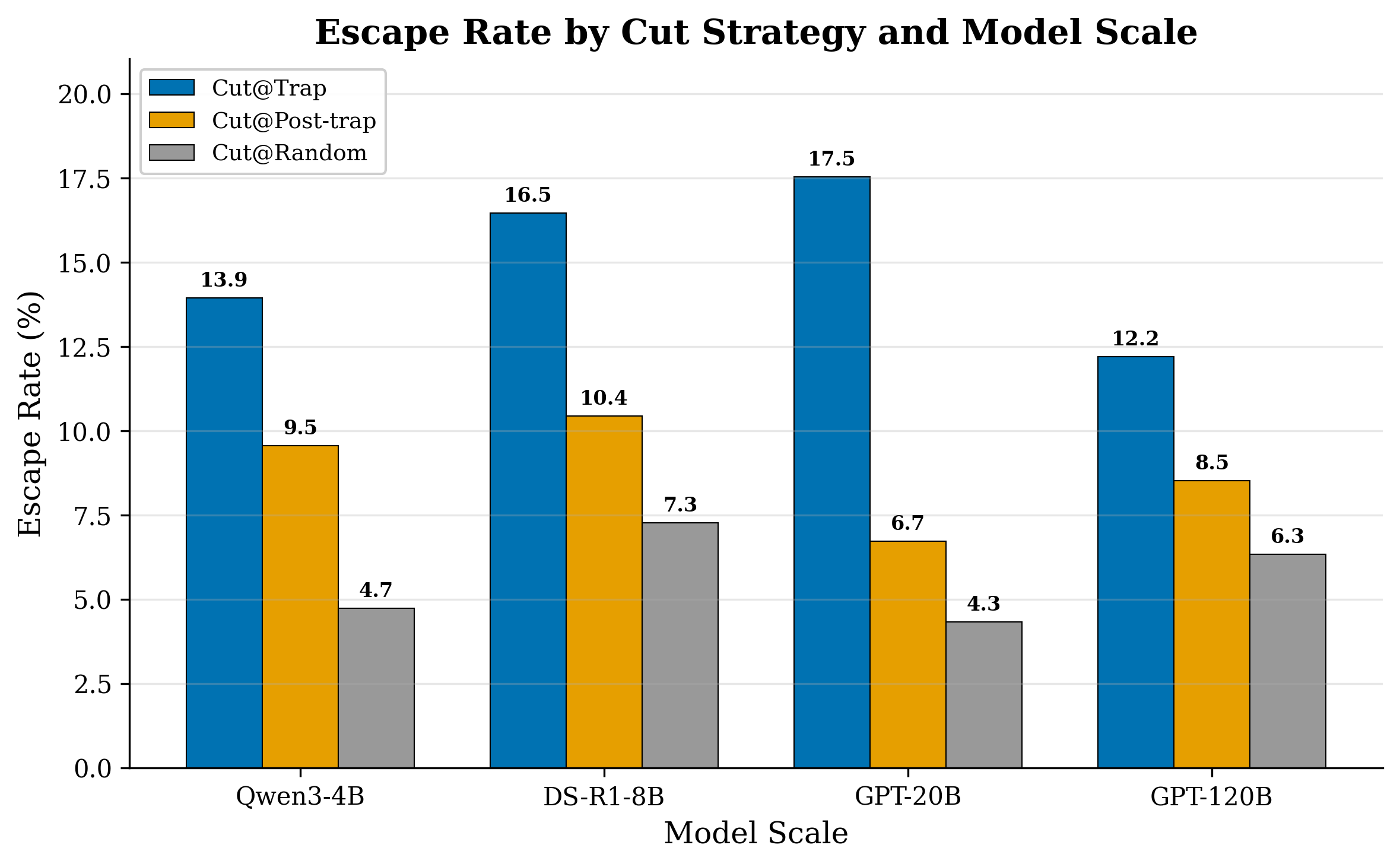}
        \caption{Escape rate by cut strategy. Truncating at the trap segment (Cut@Trap) achieves significantly higher escape rates than keeping the trap and attempting downstream correction (Cut@Post-trap).}
        \label{fig:cutpoint_comparison}
    \end{minipage}
    \hfill
    \begin{minipage}[t]{0.48\textwidth}
        \centering
        \includegraphics[width=\linewidth]{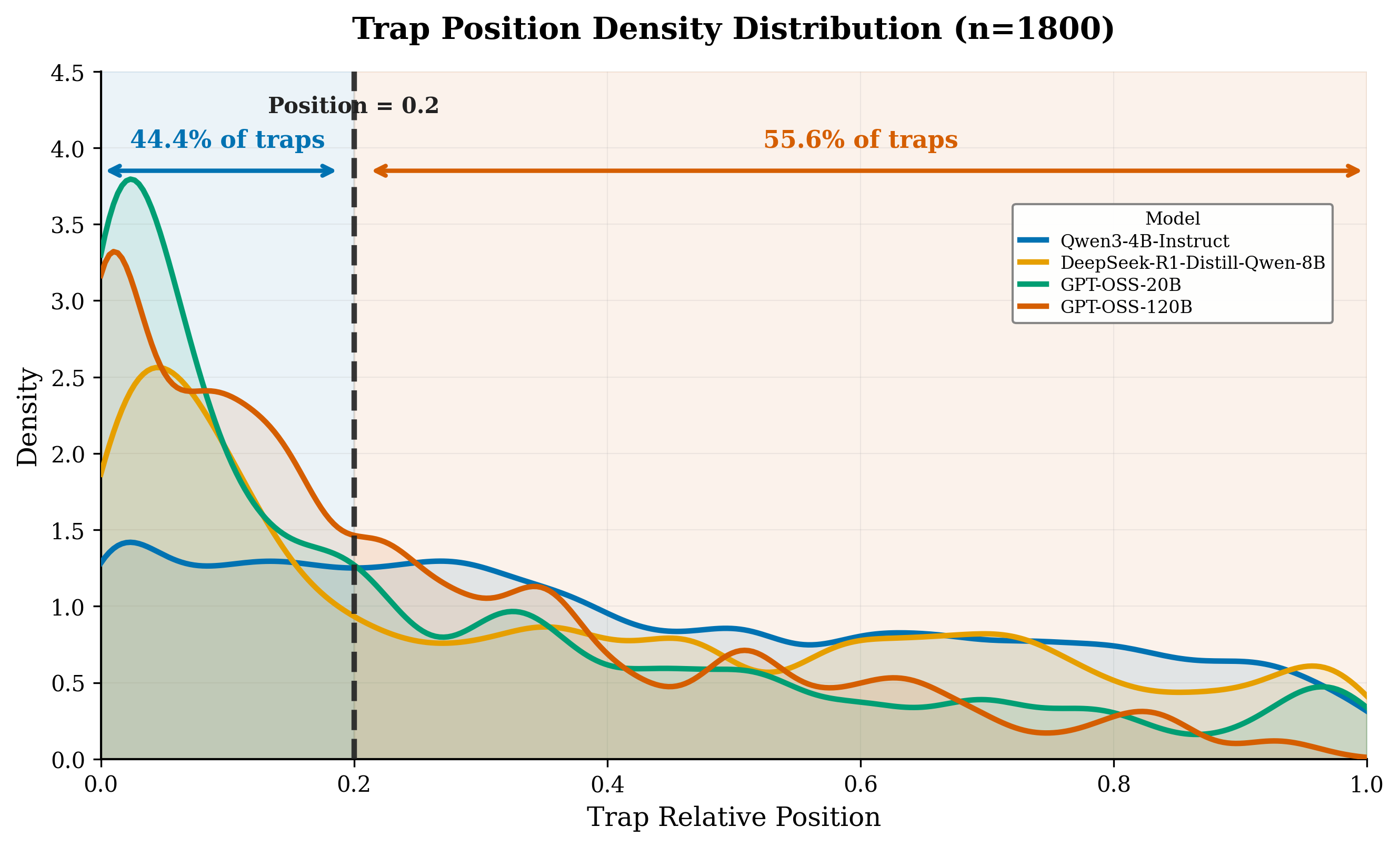}
        \caption{Trap position distribution. Traps concentrate in the early portion of trajectories, with 44.4\% occurring before relative position 0.2.}
        \label{fig:trap_position}
    \end{minipage}
\end{figure*}

Figure~\ref{fig:cutpoint_comparison} shows that Cut@Trap consistently yields the highest escape rate across all model scales. For instance, on 20B model, restarting at the trap segment achieves an escape rate of 17.5\%, while restarting from post-trap windows achieves only 6.7\% (and 4.3\% for random cuts). Similar gaps hold for 4B model (13.9\% vs.\ 9.5\% vs.\ 4.7\%) and 8B model (16.5\% vs.\ 10.4\% vs.\ 7.3\%). This indicates that once a wrong commitment is made, subsequent reasoning is often prefix-dominant and self-consistent around the error, making downstream ``fixes'' largely ineffective.

We further observe that traps concentrate in the early portion of trajectories (Figure~\ref{fig:trap_position}), with 44.4\% occurring before relative position 0.2. Together, these results validate TAAR's localization-first intervention: truncate the trap itself, rather than relying on late-stage correction attempts.

\subsection{RQ2: How Early Can We Diagnose Traps for Control?}
\label{subsec:rq2_predict}



\begin{figure*}[t]
    \centering
    
    \begin{minipage}[b]{0.48\textwidth}
        \centering
        \small
        \resizebox{\linewidth}{!}{%
            \begin{tabular}{ll|ccc}
            \toprule
            Model & Dataset & Prefix@20\% & Prefix@80\% & Full \\
            \midrule
            \multirow{4}{*}{4B}
             & AIME24 & 60.83 & 63.33 & \textbf{64.2}\\
             & AIME25 & 41.67 & \textbf{46.67}& 44.2\\
             & BRUMO25 & 54.17 & \textbf{53.33}& \textbf{55.0}\\
             & GPQA & 59.85 & 60.73 & \textbf{62.0}\\
            \midrule
            \multirow{4}{*}{20B}
             & AIME24 & 74.17 & 75.83 & \textbf{78.3}\\
             & AIME25 & 76.67 & 75.00 & \textbf{77.5}\\
             & BRUMO25 & 79.17 & 81.67 & \textbf{80.8}\\
             & GPQA & \textbf{60.61}& \textbf{60.61}& 59.0 \\
            \bottomrule
            \end{tabular}%
        }
        
        \vspace{-5pt} 
        
        \captionof{table}{Early diagnosis efficiency. Prefix@X\%: the diagnostic policy receives only the first X\% of the trajectory. TAAR achieves comparable performance even with partial observations.}
        \label{tab:prefix_scaling}
    \end{minipage}
    \hfill 
    \begin{minipage}[b]{0.48\textwidth}
        \centering
        \includegraphics[width=\linewidth]{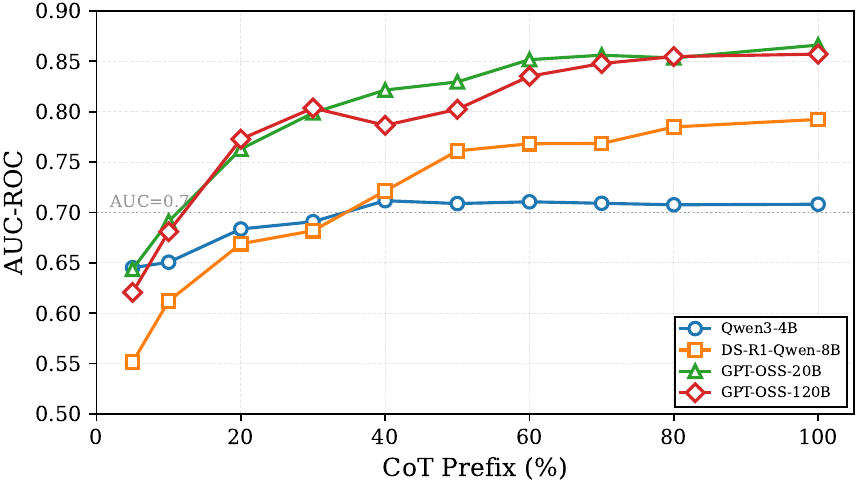}
        
        \vspace{-5pt}
        
        \captionof{figure}{Trap detection AUC-ROC vs. CoT prefix length. Detection exceeds 0.7 AUC with only 20\% of the trajectory and saturates around 40--60\%.}
        \label{fig:early_diagnosis}
    \end{minipage}
    
    \vspace{-10pt}
\end{figure*}

We further test early diagnosis by providing only a prefix of the reasoning trajectory to the policy (Table~\ref{tab:prefix_scaling}). TAAR retains comparable downstream accuracy even with partial observations. For example, for 4B model on AIME24, performance increases from 60.83 (Prefix@20\%) to 64.2 (Full), and for 20B model on BRUMO25, performance is already strong at Prefix@20\% (79.17) and remains comparable at Full (80.8).



Figure~\ref{fig:early_diagnosis} corroborates this trend: trap detection AUC-ROC exceeds 0.7 with only 20\% of the trajectory and saturates around 40--60\%, suggesting that diagnosis can be performed online without waiting for full generation. This is a key practical advantage: TAAR does not require completing the entire long chain before deciding whether to intervene.

\subsection{RQ3: Does Adaptive Cut Position Outperform Fixed Heuristics?}
\label{subsec:rq3_restart_index}

A natural question is whether content-aware cut prediction provides value over simple fixed heuristics. We compare against fixed-position baselines that truncate the trajectory at predetermined relative positions (25\%, 50\%, 75\%), regardless of reasoning content.


Table~\ref{tab:idx_control} shows that \textbf{no single fixed cut position works consistently across datasets}. For the 4B model, fixed cuts yield average accuracies of 54.1 (25\%), 52.7 (50\%), and 53.2 (75\%), whereas TAAR achieves 56.4. For the 20B model, fixed cuts range from 72.6 to 73.1 on average, while TAAR reaches 73.9. The variability arises because trap location is instance-dependent: some problems go wrong early, others mid-trajectory. Adaptive localization is necessary to robustly remove the erroneous commitment without excessive truncation.

This motivates the next question: given a predicted cut position, can an additional control signal decide \emph{when} to restart and \emph{how strongly} to perturb decoding? We address this in RQ4 using the escape probability $\hat{p}$.

\subsection{RQ4: Does Escape Probability Improve Control?}
\label{subsec:rq4_adaptive}

Beyond trap localization, does the predicted escape probability $\hat{p}$ provide additional control benefit? We find that $\hat{p}$ is discriminative of trajectory correctness (see Appendix~\ref{app:escape_dist} for distribution analysis), motivating its use for adaptive intervention.

To isolate the contribution of $\hat{p}$, we compare three variants: (i)~\textbf{Cut@AllTraps}, which restarts at every detected trap regardless of $\hat{p}$; (ii)~\textbf{Random $\hat{p}$}, which uses TAAR's predicted cut position but randomizes the escape probability to remove the gating effect; and (iii)~\textbf{TAAR}, which uses both predicted position and $\hat{p}$ to adaptively select intervention strength.


\begin{table}[H]
    \centering
    
    \begin{minipage}[c]{0.48\textwidth} 
        \centering
        \small
        \resizebox{\linewidth}{!}{%
            \begin{tabular}{l|cccc|c}
            \toprule
            Strategy & AIME24 & AIME25 & BRUMO25 & GPQA & Avg. \\
            \midrule
            \multicolumn{6}{c}{\textit{4B Model}} \\
            \midrule
            Cut@25\% & 59.2 & 43.3 & 53.3 & 60.7 & 54.1 \\
            Cut@50\% & 60.8 & 36.7 & 53.3 & 60.1 & 52.7 \\
            Cut@75\% & 58.3 & 39.2 & 54.2 & 61.0 & 53.2 \\
            TAAR (Ours) & \textbf{64.2} & \textbf{44.2} & \textbf{55.0} & \textbf{62.0} & \textbf{56.4} \\
            \midrule
            \multicolumn{6}{c}{\textit{20B Model}} \\
            \midrule
            Cut@25\% & 74.2 & 75.8 & 79.2 & \textbf{61.1} & 72.6 \\
            Cut@50\% & 76.7 & 75.8 & 77.5 & 60.4 & 72.6 \\
            Cut@75\% & 75.8 & 76.7 & 80.0 & 59.7 & 73.1 \\
            TAAR (Ours) & \textbf{78.3} & \textbf{77.5} & \textbf{80.8} & 59.0 & \textbf{73.9} \\
            \bottomrule
            \end{tabular}%
        }
        \vspace{2pt} 
        \caption{Cut position ablation on AIME24/AIME25/BRUMO25/GPQA. Fixed-position cuts show inconsistent results, while TAAR's adaptive prediction achieves the best average performance.}
        \label{tab:idx_control}
    \end{minipage}
    \hfill 
    \begin{minipage}[c]{0.48\textwidth}
        \centering
        \small
        \resizebox{\linewidth}{!}{%
            \begin{tabular}{l|cccc|c}
            \toprule
            Strategy & AIME24 & AIME25 & BRUMO25 & GPQA & Avg. \\
            \midrule
            \multicolumn{6}{c}{\textit{4B Model}} \\
            \midrule
            Cut@AllTraps & 65.0 & 42.5 & 55.0 & 57.1 & 54.9 \\
            Random $\hat{p}$ & 64.2 & 35.8 & 55.8 & 60.9 & 54.2 \\
            TAAR (Ours) & \textbf{64.2} & \textbf{44.2} & \textbf{55.0} & \textbf{62.0} & \textbf{56.4} \\
            \midrule
            \multicolumn{6}{c}{\textit{20B Model}} \\
            \midrule
            Cut@AllTraps & 67.5 & 76.7 & 76.7 & 55.3 & 69.1 \\
            Random $\hat{p}$ & 69.2 & 79.2 & 80.0 & 59.1 & 71.9 \\
            TAAR (Ours) & \textbf{78.3} & \textbf{77.5} & \textbf{80.8} & \textbf{59.0} & \textbf{73.9} \\
            \bottomrule
            \end{tabular}%
        }
        \vspace{2pt}
        \caption{Escape probability ablation. Cut@AllTraps ignores $\hat{p}$ entirely; Random $\hat{p}$ uses predicted position but randomizes escape probability. TAAR's predicted $\hat{p}$ improves control decisions.}
        \label{tab:escape_control}
    \end{minipage}
\end{table}

As shown in Table~\ref{tab:escape_control}, $\hat{p}$ improves control decisions. On the 20B model, TAAR achieves the best average accuracy (73.9), outperforming Random $\hat{p}$ (71.9) and Cut@AllTraps (69.1). This indicates that \textbf{not all detected traps warrant the same intervention}: some trajectories can self-repair with continuation or mild resampling, while severely trapped cases benefit from stronger restarts.

\paragraph{Computational efficiency.}
Escape probability also improves computational efficiency via adaptive gating. The ``Baseline'' column in Table~\ref{tab:token_efficiency} shows total tokens for 4-sample averaging; ``TAAR'' and ``Ablation'' columns show \emph{extra tokens beyond baseline} incurred by each method.

\begin{table}[t]
\centering
\small
\resizebox{\columnwidth}{!}{%
\begin{tabular}{l|r|rr|rr|r}
\toprule
\textbf{Model} & \textbf{Baseline (Avg@4)} & \textbf{Extra (TAAR)} & \textbf{Extra/Base} & \textbf{Extra (Ablation)} & \textbf{Extra/Base} & \textbf{Savings} \\
\midrule
4B & 1,735,110 & 576,918 & 33.2\% & 1,377,418 & 79.4\% & 58.1\% \\
8B & 4,388,791 & 1,279,756 & 29.2\% & 2,353,417 & 53.6\% & 45.6\% \\
20B & 5,617,370 & 2,190,737 & 39.0\% & 2,725,979 & 48.5\% & 19.6\% \\
120B & 3,500,167 & 704,376 & 20.1\% & 1,818,700 & 52.0\% & 61.3\% \\
\midrule
Total & 15,241,438 & 4,751,787 & 31.2\% & 8,275,514 & 54.3\% & 42.6\% \\
\bottomrule
\end{tabular}%
}
\caption{Token efficiency. ``Extra'' columns show additional tokens beyond baseline incurred by each method. TAAR incurs 31.2\% extra tokens relative to baseline, while the ablation (Cut@AllTraps) incurs 54.3\%, yielding 42.6\% savings.}
\label{tab:token_efficiency}
\end{table}

Table~\ref{tab:token_efficiency} shows TAAR incurs only 31.2\% additional tokens beyond baseline, compared to 54.3\% for the ablation that cuts at all detected traps. This yields a 42.6\% reduction in extra overhead. The result demonstrates that $\hat{p}$ is not only a performance signal but also a practical knob for controlling test-time cost. This supports our thesis that test-time scaling should be viewed as controlling effective prefix, not merely generating longer continuations.

\subsection{RQ5: What Changes After Restart, and Why Does It Work?}
\label{subsec:rq5_mechanism}

We analyze qualitative reasoning dynamics before and after restart to understand why TAAR improves performance. Across cases, a common pattern emerges: once an early wrong commitment enters the context, later deliberation tends to remain prefix-dominant and rationalize the error. TAAR breaks this deadlock by truncating the corrupted prefix and prompting a counterfactual re-derivation.

We analyze qualitative reasoning dynamics before and after restart to understand why TAAR improves performance. A recurring pattern in trapped trajectories is that once an early wrong commitment enters the context, later deliberation becomes prefix-dominant and tends to rationalize the initial error rather than revise it. TAAR breaks this deadlock by truncating the corrupted prefix and prompting a counterfactual re-derivation from the remaining clean context.

\begin{wrapfigure}{r}{0.5\columnwidth}
    \vspace{-20pt} 
    
    \centering
    \includegraphics[width=\linewidth]{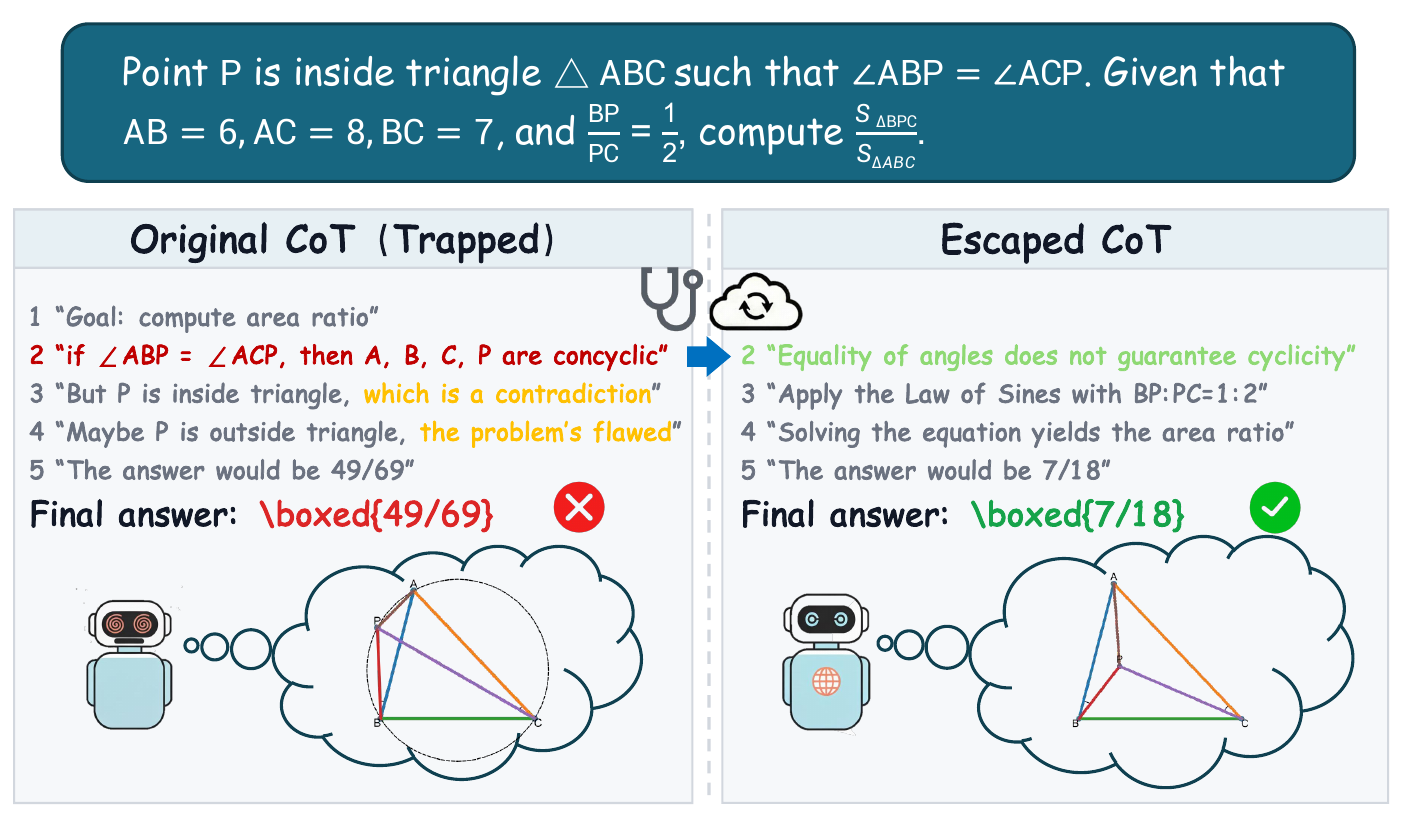}
    
    \caption{Geometry case study illustrating a prefix-dominant trap and how TAAR restarts by cutting before the wrong commitment.}
    \label{fig:case_study_geom}
    
    \vspace{-10pt} 
\end{wrapfigure}

\paragraph{Case study.}
Figure~\ref{fig:case_study_geom} shows a representative geometry problem where the model incorrectly assumes a concyclic polygon configuration at segment 12. The diagnostic policy predicts $\hat{t}=11$, truncating the trajectory immediately before the erroneous assumption is introduced. After restart, the model re-derives the configuration without imposing concyclicity, identifies the correct irregular structure, and reaches the correct answer.

This case illustrates that restart is effective not because it adds more downstream computation, but because it changes the effective prefix that conditions generation, enabling the model to explore a different reasoning branch. In practice, TAAR shortens the verify--correct loop by preventing prolonged self-consistent but incorrect continuations that are anchored to an early mistaken premise.

%% file: section/7_discussion/discussion.tex
\section{Discussion and Conclusion}

\paragraph{TAAR as test-time diagnostic control.}
Our results suggest that many Long-CoT failures are not caused by insufficient test-time compute, but by misallocated compute after an early wrong commitment enters the context. TAAR addresses this by predicting two control-relevant signals from partial trajectories: a trap index indicating where to remove the corrupted prefix, and an escape probability indicating whether and how strongly to restart. Under a fixed sampling budget, this simple truncation-and-adaptive-restart mechanism reallocates compute toward counterfactual re-derivations rather than extending trapped continuations.

\paragraph{When does TAAR help, and how does it relate to other methods?}
TAAR tends to help most on harder benchmarks and for small-to-mid scale reasoners, where prefix-dominant traps are common and restarting can substantially change the explored solution path. For very strong models, gains may be smaller and less consistent under small budgets, since many prefixes are already near-correct and aggressive perturbations can discard useful work. TAAR is complementary to outcome-based selection and structured search: it aims to change the candidate distribution by removing wrong commitments, after which verifiers or ranking can be applied for final selection.

In summary, we formalize Thinking Traps as prefix-dominant deadlocks and show that effective recovery often requires modifying the effective prefix. TAAR provides a lightweight test-time controller that localizes traps and adaptively restarts decoding, improving reasoning performance under a fixed compute budget without updating base model parameters.

%% file: section/8_limitations/limitation.tex
\section{Limitations}

TAAR has several limitations. First, trap localization operates over paragraph-based segments, introducing boundary ambiguity that can lead to over- or under-truncation when wrong commitments span multiple segments. Second, our offline supervision relies on an LLM judge with limited manual auditing, so label noise and judge-specific biases may propagate into the diagnostic policy. Third, escape probability estimation depends on automatic verifiers, which works well for math but may not transfer to open-ended tasks where correctness is subjective. Fourth, for very strong models where prefixes are often near-correct, aggressive restarts can discard useful work and reduce net gains. Finally, TAAR assumes access to explicit Long-CoT traces that can be segmented and truncated, which may not hold for systems with hidden reasoning or constrained APIs.

%% file: section/9_appendix/appendix.tex
\appendix
\section{Segmentation Function}
\label{app:segmentation}
We extract the reasoning portion of each output (e.g., the content inside \texttt{<think>...</think>} when available) and segment it into $T$ continuous segments using paragraph boundaries (\verb|\n\n|).
To stabilize index-based localization, we enforce a minimum segment length 200 characters by merging short chunks with adjacent segments.

\section{LLM Judge Prompt for Trap and Window Annotation}
\label{app:glm_prompt}

The following is the prompt template used with GLM-4.7 to annotate trap indices, escape points, and self-repair windows.

\begin{tcolorbox}[colback=gray!10, colframe=gray!50, boxrule=0.5pt, breakable]
\small
\begin{verbatim}
You are a "long-CoT reasoning trap locator."

[Problem]
{problem}

[Input]
A long CoT text segmented with labels:
{segmented_cot}

[Ground Truth Answer]
{ground_truth}

[Task]
1. Identify exactly one trap in the CoT text:
   a. A "trap" is the earliest critical erroneous assumption,
      unjustified leap, or improper simplification that "locks"
      or severely restricts subsequent reasoning.
   b. Consequence: subsequent reasoning space becomes
      significantly constrained, leading to failure or deviation.
   c. If multiple candidates exist, select the earliest and
      most restrictive one.

2. In the entire text, find segments directly related to the
   identified trap (only output labels without repeating their
   contents):

   High-precision eligibility (MUST satisfy; otherwise exclude):
   A segment is eligible ONLY IF it explicitly contains
   meta-reasoning cues targeting the trap, i.e. it explicitly
   does at least one of:
   - Reflection points: explicitly doubt/question the trap
     assumption itself OR a direct consequence of it, but fail
     to correct it.
   - New approach points: explicitly propose a different
     method/representation/strategy to escape, but still rely
     on the trap assumption (do not fix it).
   - Verification points: explicitly check the trap assumption
     OR a direct consequence via examples/boundaries/calculations,
     but miss the key flaw.

   NOT eligible: segments that merely continue routine
   computation/derivation along the trapped path WITHOUT explicit
   doubt / alternative attempt / verification.

   Relevance ranking (internal; do NOT output scores):
   - For each eligible candidate, assign rel in {3,2,1}:
     rel=3: explicitly target the trap assumption itself
            (name/restatement/check) OR explicitly attempt
            to escape it.
     rel=2: explicitly target a direct consequence that
            critically depends on the trap, with
            doubt/alternative/check.
     rel=1: weak/implicit relation -> EXCLUDE (do not output).
   - Keep ONLY candidates with rel >= 2.
   - Each list must be sorted by (rel descending, index ascending).
     Output labels only.

   Selection constraints (precision-first):
   - Do NOT include the trap segment itself; all points must
     satisfy index > trap index.
   - No duplicates; a label can appear in at most ONE list.
   - If a segment fits multiple categories, assign it to the
     most specific with priority:
     new_approach_points > verification_points > reflection_points.
   - Hard caps (no total cap): reflection_points <= 3,
     new_approach_points <= 3, verification_points <= 3.
   - (These arrays may be empty; it is OK to output [].)

3. Determine if escaped:
   a. If any later segment explicitly corrects the trap assumption
      and breaks free from the erroneous path, set
      trap_type="escaped successfully" and "escape_point" to the
      earliest correcting segment.
   b. Otherwise, set trap_type="did not escape" and
      "escape_point"="".

[Output]
Output only valid JSON (no explanations or extra text):
{
  "trap": "cot_x" or "",
  "trap_type": "escaped successfully" or "did not escape" or "",
  "escape_point": "cot_y" or "",
  "reflection_points": ["cot_i", ...],
  "new_approach_points": ["cot_j", ...],
  "verification_points": ["cot_m", ...]
}

[Empty Output]
If no trap satisfying "maximum causal influence/strongest lock"
is found:
{
  "trap": "",
  "trap_type": "",
  "escape_point": "",
  "reflection_points": [],
  "new_approach_points": [],
  "verification_points": []
}
\end{verbatim}
\end{tcolorbox}

\section{Reboot Suffix Templates}
\label{app:reboot_suffix}

For hard restarts, we append a structured reboot suffix to the truncated prefix. The main experiments use the English suffix (same language as the prompt). We also provide multilingual variants for optional code-switch experiments.

\paragraph{English suffix (used in main experiments).} \mbox{} \\
\begin{tcolorbox}[colback=gray!10, colframe=gray!50, boxrule=0.5pt, width=\columnwidth]
    \footnotesize
    \ttfamily
    \raggedright 
    Wait, let me completely rethink this problem in English. The previous chain of thought might be limited, so I need to reorganize my thoughts in English and analyze from scratch.
\end{tcolorbox}

\paragraph{Multilingual variants.}
We also provide translated variants of the English reboot suffix in Chinese, Korean, Russian, Arabic, and French for code-switch experiments (see Appendix~\ref{app:codeswitch}).

\section{Additional Details for Escape Probability Estimation}
\label{app:escape_estimation}

We use a total resampling budget of $N=36$ per trajectory. We sample from post-trap windows
(reflection, new-approach, and verification; each capped at 3) and allocate leftover samples to random cut points
uniformly drawn from $[t^*+1, T-1]$. Decoding uses temperature=0.7, top-$p$ disabled, and total context+generation length $\leq$32k tokens.
Correctness is evaluated by the math-verify verifier.

\section{Trap Position and Escape Rate Analysis}
\label{app:position_escape}
\begin{figure}[t]
    \centering
    \small
    \includegraphics[width=0.5\columnwidth]{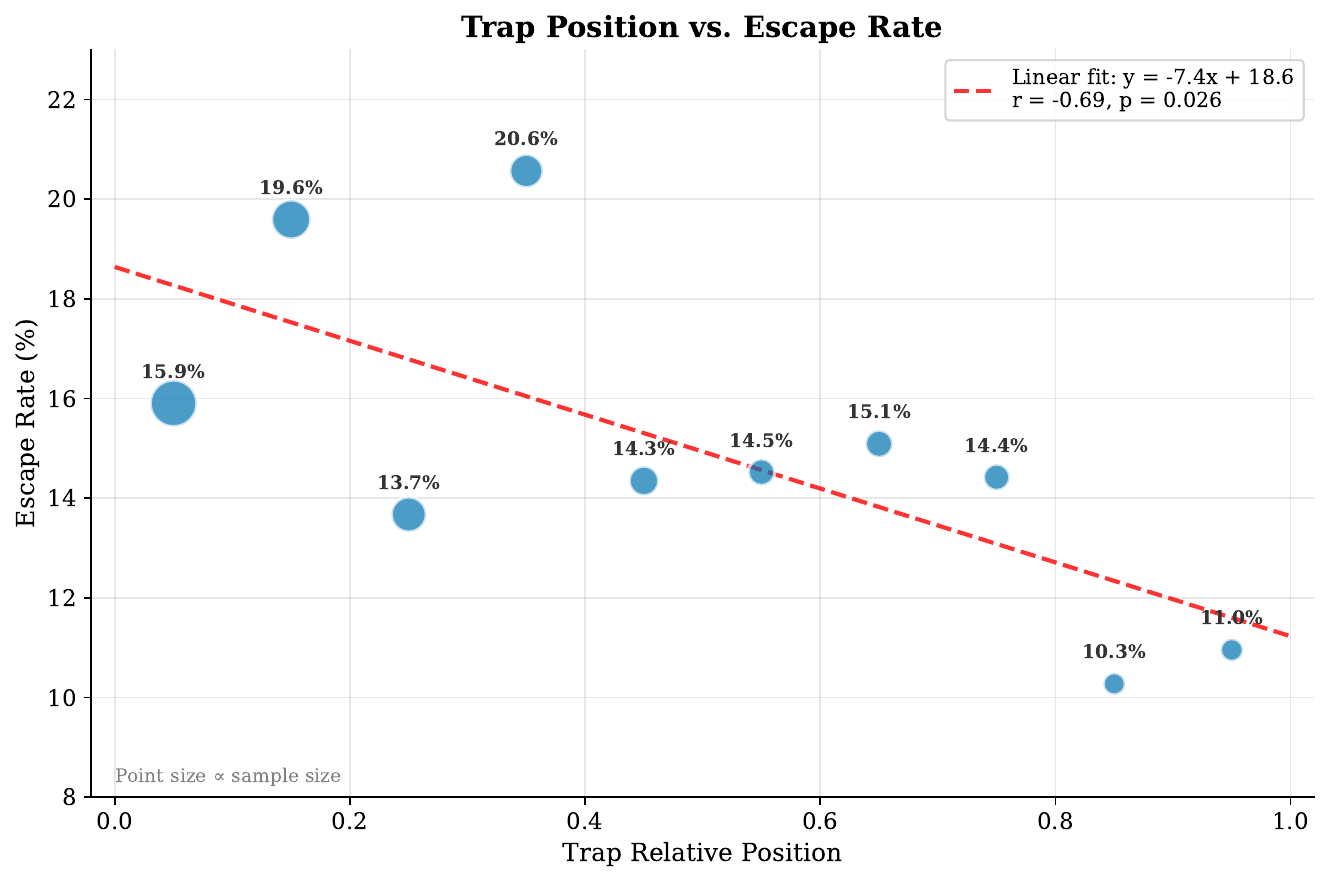}
    \caption{Relationship between trap relative position and escape rate on a subset of the offline data construction dataset (Section~\ref{sec:data_construction}). Earlier traps exhibit higher escape rates, while later traps are harder to recover from. The negative correlation ($r=-0.69$, $p=0.026$) suggests that early wrong commitments leave more room for self-correction, whereas late traps constrain the remaining trajectory too severely.}
    \label{fig:position_escape}
\end{figure}

Figure~\ref{fig:position_escape} shows the relationship between trap position and empirical escape rate on a subset of our constructed dataset. We observe a significant negative correlation: traps occurring in the early portion of trajectories (relative position $<0.2$) have escape rates around 16--20\%, while traps in the late portion (relative position $>0.8$) drop to 10--11\%. Earlier traps leave more remaining trajectory for the model to reflect, reconsider, and potentially self-correct; later traps leave little room for recovery before the final answer.

\section{Filtering Rules and Loss Breakdown}
\label{app:filtering_details}

\paragraph{Preprocess filtering.}
We remove records with API call errors, invalid JSON, or missing required fields.
Across all raw records (6,000):
\begin{itemize}
    \item API errors: 1,367 (22.4\%)
    \item JSON parse errors: 52 (0.9\%)
\end{itemize}

\paragraph{Consistency filtering.}
From 4,680 preprocessed records, we remove:
\begin{itemize}
    \item (i) trap + did-not-escape + correct: 406 (8.7\%)
    \item (ii) trap + escaped + incorrect: 343 (7.3\%)
    \item (iii) no-trap + incorrect: 270 (5.8\%)
\end{itemize}
Total removed by consistency filtering: 1,019 (21.8\%).

\section{Pattern-Based Difficulty and Dataset Composition}
\label{app:pattern_stats}

We define a 4-bit correctness pattern [120B][20B][8B][4B], where 1 indicates correct final answer and 0 indicates incorrect.
We bucket problems by difficulty level based on the number of correct models.

\begin{table}[h]
\centering
\small
\begin{tabular}{lrr}
\toprule
Difficulty level & \# trajectories & Trap rate \\
\midrule
1 (3/4 correct) & 591 & 53.0\% \\
2 (2/4 correct) & 1,394 & 73.5\% \\
3 (1/4 correct) & 1,125 & 89.0\% \\
4 (0/4 correct) & 551 & 100.0\% \\
\bottomrule
\end{tabular}
\caption{Difficulty distribution and trap rates in the final dataset ($n$=3,661).}
\label{tab:difficulty_traprate}
\end{table}

\begin{table}[h]
\centering
\small
\begin{tabular}{lrrrr}
\toprule
Pattern & Total & With trap & No trap & Trap rate \\
\midrule
1100 & 699 & 483 & 216 & 69.1\% \\
1000 & 645 & 583 & 62  & 90.4\% \\
0000 & 551 & 551 & 0   & 100.0\% \\
1110 & 421 & 231 & 190 & 54.9\% \\
1010 & 367 & 274 & 93  & 74.7\% \\
1001 & 191 & 161 & 30  & 84.3\% \\
0010 & 173 & 134 & 39  & 77.5\% \\
1101 & 170 & 82  & 88  & 48.2\% \\
0001 & 169 & 161 & 8   & 95.3\% \\
0100 & 138 & 123 & 15  & 89.1\% \\
0110 & 88  & 72  & 16  & 81.8\% \\
0101 & 49  & 35  & 14  & 71.4\% \\
\bottomrule
\end{tabular}
\caption{Pattern distribution in the final dataset ($n$=3,661).}
\label{tab:pattern_dist_final}
\end{table}

\section{Train/Dev/Test Split}
\label{app:split}

We split problems into Train/Dev/Test with a ratio of 80/10/10 using a fixed random seed (42).
For each problem, models at different scales (4B, 8B, 20B, and 120B) generate independent reasoning trajectories; all such trajectories are assigned to the same split as the underlying problem, ensuring no cross-split leakage across model sizes.
Table~\ref{tab:pattern_split} reports the per-pattern split counts.

\begin{table}[h]
\centering
\small
\begin{tabular}{lrrrr}
\toprule
Pattern & Train & Dev & Test & Total \\
\midrule
1110 & 116 & 14 & 10 & 140 \\
1101 & 48  & 6  & 6  & 60  \\
1100 & 189 & 14 & 32 & 235 \\
1010 & 105 & 17 & 11 & 133 \\
1001 & 70  & 5  & 2  & 77  \\
0110 & 33  & 0  & 2  & 35  \\
0101 & 17  & 3  & 0  & 20  \\
1000 & 185 & 26 & 21 & 232 \\
0100 & 47  & 2  & 6  & 55  \\
0010 & 68  & 12 & 10 & 90  \\
0001 & 60  & 9  & 4  & 73  \\
0000 & 262 & 42 & 46 & 350 \\
\midrule
Total & 1200 & 150 & 150 & 1500 \\
\bottomrule
\end{tabular}
\caption{Problem split by pattern (1,500 problems).}
\label{tab:pattern_split}
\end{table}

\section{Manual Audit Protocol}
\label{app:audit}

We randomly sample 100 instances for manual re-check of trap index and window eligibility to sanity-check LLM annotations.

\paragraph{Audit procedure.}
Two annotators independently reviewed each sampled instance, checking:
\begin{itemize}
    \item Whether the identified trap segment is indeed the earliest wrong commitment
    \item Whether the self-repair windows are correctly classified (reflection, new-approach, or verification)
    \item Whether the escape point (if any) is correctly identified
\end{itemize}

\section{Diagnostic Policy Input/Output Templates}
\label{app:templates}

This section describes the formatting templates $\mathcal{T}_{\text{in}}(\cdot)$ and $\mathcal{T}_{\text{out}}(\cdot)$ used to construct training data for the diagnostic policy $\pi_\phi$.

\subsection{Input Template \texorpdfstring{$\mathcal{T}_{\text{in}}$}{T\_in}}

The input to the diagnostic policy concatenates the model identifier, problem statement, and the segmented reasoning trace with explicit segment labels:

\begin{tcolorbox}[colback=gray!10, colframe=gray!50, boxrule=0.5pt]
\small
\begin{verbatim}
Please identify and locate the trap in the current problem's
reasoning process, and provide the escape action.

[Model]
{model_name}

[Problem]
{problem_statement}

[Reasoning Process]
<cot_0>
{segment_0_text}

<cot_1>
{segment_1_text}

...

<cot_K>
{segment_K_text}

Output your analysis in JSON format:
\end{verbatim}
\end{tcolorbox}
\subsection{Output Template \texorpdfstring{$\mathcal{T}_{\text{out}}$}{T\_out}}

The output format encodes the trap index and escape probability in JSON:

\begin{tcolorbox}[colback=gray!10, colframe=gray!50, boxrule=0.5pt]
\small
\begin{verbatim}
{
  "trap_index": "t*",
  "escape_probability": "p_escape",
  "extra": {extra information}
}
\end{verbatim}
\end{tcolorbox}

During training, we provide gold labels $(t^*, p_{\text{escape}})$ from the offline annotation pipeline.
During inference, the policy outputs predictions $(\hat{t}, \hat{p})$ which are used by the adaptive restart controller.

\section{Diagnostic Policy Evaluation Details}
\label{app:policy_eval}

We evaluate the diagnostic policy on test samples spanning four model scales. This section provides detailed breakdowns beyond the summary in Section~\ref{subsec:rq2_predict}.

\paragraph{Trap detection by model scale.}
Table~\ref{tab:detection_by_model} shows that detection rates vary across model scales. The policy achieves near-perfect detection on 4B trajectories (98.9\%) but lower rates on larger models (55--67\%). This is expected: larger models produce more subtle errors that are harder for an external diagnostic model to identify.

\begin{table*}[t]
    \centering
    
    \begin{minipage}[c]{0.48\linewidth} 
        \centering
        \small
        \begin{tabular}{lrrr}
            \toprule
            Model & Total & Detected & Rate \\
            \midrule
            4B & 180 & 178 & 98.9\% \\
            8B & 45 & 30 & 66.7\% \\
            20B & 43 & 24 & 55.8\% \\
            120B & 90 & 50 & 55.6\% \\
            \midrule
            Overall & 358 & 282 & 78.8\% \\
            \bottomrule
        \end{tabular}
        \vspace{5pt}
        \caption{Trap detection rate by model scale. Detection is easier for smaller models whose errors tend to be more explicit.}
        \label{tab:detection_by_model}
    \end{minipage}
    \hfill 
    \begin{minipage}[c]{0.48\linewidth}
        \centering
        \small
        \begin{tabular}{lr}
            \toprule
            Metric & Value \\
            \midrule
            Top-1 Accuracy & 29.1\% \\
            Within $\pm 3$ & 55.3\% \\
            Mean $|\hat{t} - t^*|$ & 9.46 \\
            Avg. CoT segments & 55.6 \\
            Relative Error & 17.0\% \\
            \bottomrule
        \end{tabular}
        \vspace{5pt}
        \caption{Overall localization accuracy on detected traps.}
        \label{tab:localization_overall}
    \end{minipage}
\end{table*}

\paragraph{Localization accuracy.}
Among detected traps, Table~\ref{tab:localization_overall} reports position prediction accuracy. While Top-1 exact match is modest (29.1\%), the mean absolute error of 9.46 segments represents only 17.0\% of the average trajectory length (55.6 segments), and Within $\pm 3$ reaches 55.3\%.




\paragraph{Localization by distance to truncation point.}
Table~\ref{tab:loc_by_distance} shows that localization accuracy degrades as the trap occurs further from the truncation point. When the trap is within 1 step of truncation, Top-1 reaches 62.3\%; when it is more than 20 steps away, Top-1 drops to 9.7\%. This motivates early diagnosis: the sooner we detect a trap after it occurs, the more accurately we can localize it.



\begin{table*}[t]
    \centering
    
    \begin{minipage}[c]{0.48\linewidth}
        \centering
        \small
        \resizebox{\linewidth}{!}{%
            \begin{tabular}{lrrrr}
            \toprule
            Distance & N & Top-1 & Within $\pm 3$ & $|\hat{t} - t^*|$ \\
            \midrule
            1 step & 53 & 62.3\% & 77.4\% & 4.30 \\
            2--3 steps & 33 & 27.3\% & 84.8\% & 3.27 \\
            4--5 steps & 22 & 40.9\% & 72.7\% & 10.86 \\
            6--10 steps & 53 & 18.9\% & 56.6\% & 5.53 \\
            11--20 steps & 49 & 28.6\% & 44.9\% & 10.90 \\
            $>$20 steps & 72 & 9.7\% & 26.4\% & 17.57 \\
            \bottomrule
            \end{tabular}%
        }
        
        \vspace{5pt}
        
        \caption{Localization accuracy by distance from trap to truncation point.}
        \label{tab:loc_by_distance}
    \end{minipage}
    \hfill 
    \begin{minipage}[c]{0.48\linewidth}
        \centering
        \small
        \resizebox{\linewidth}{!}{%
            \begin{tabular}{lrrrr}
            \toprule
            Input Length & N & Top-1 & Within $\pm 3$ & $|\hat{t} - t^*|$ \\
            \midrule
            $\leq$10 & 59 & 66.1\% & 93.2\% & 0.69 \\
            11--20 & 48 & 41.7\% & 68.8\% & 3.31 \\
            21--40 & 75 & 22.7\% & 52.0\% & 7.12 \\
            41--60 & 45 & 6.7\% & 24.4\% & 13.16 \\
            $>$60 & 55 & 5.5\% & 32.7\% & 24.38 \\
            \bottomrule
            \end{tabular}%
        }
        
        \vspace{5pt}
        
        \caption{Localization accuracy by input sequence length (number of segments).}
        \label{tab:loc_by_length}
    \end{minipage}
\end{table*}

\paragraph{Localization by input length.}
Table~\ref{tab:loc_by_length} shows that shorter input sequences yield better localization. For sequences with $\leq$10 segments, Top-1 reaches 66.1\% and Within $\pm 3$ reaches 93.2\%. Performance degrades for longer sequences due to increased search space and noise.

\begin{table}[h]
\centering
\small
\begin{tabular}{lr}
\toprule
Metric & Value \\
\midrule
Correlation & 0.336 \\
\bottomrule
\end{tabular}
\caption{Escape probability prediction accuracy.}
\label{tab:escape_pred_detail}
\end{table}

\paragraph{Escape probability prediction.}
Table~\ref{tab:escape_pred_detail} reports the accuracy of escape probability predictions. The positive correlation ($r=0.336$) indicate that predicted $\hat{p}$ provides a useful signal for adaptive gating.

\label{app:codeswitch}

\begin{table}[h]
\centering
\small
\setlength{\tabcolsep}{3.5pt}  
\begin{tabular}{@{}llccccc@{}}
\toprule
\textbf{Mdl} & \textbf{Data} & \textbf{Mono} & \textbf{Sw-ar} & \textbf{Sw-fr} & \textbf{Sw-zh} & \textbf{Hi-T} \\
\midrule
\multirow{3}{*}{4B}  
  & AIME24   & 64.2 & 60.0 & 59.2 & 60.0 & 65.7 \\
  & AIME25   & 44.2 & 43.3 & 45.0 & 43.3 & 44.2 \\
  & BRUMO25  & 55.0 & 53.3 & 54.2 & 54.2 & 53.3 \\
\midrule
\multirow{3}{*}{20B} 
  & AIME24   & 78.3 & 77.5 & 76.7 & 77.5 & 75.0 \\
  & AIME25   & 77.5 & 76.7 & 79.2 & 78.3 & 80.0 \\
  & BRUMO25  & 80.0 & 81.7 & 83.3 & 80.8 & 85.5 \\
\bottomrule
\end{tabular}
\caption{Comparison of hard restart strategies across multilingual perturbations. Mono: monolingual reboot suffix; Sw-ar/fr/zh: code-switching to Arabic, French, or Chinese; Hi-T: high-temperature resampling. No single strategy consistently dominates}
\label{tab:codeswitch}
\end{table}

\section{Code-Switch Experiments}
We conduct exploratory experiments using multilingual reboot suffixes (code-switching) as an alternative hard restart operator. The hypothesis is that switching the reasoning language may provide a stronger perturbation to break free from prefix-dominant traps.

\paragraph{Setup.}
For trajectories flagged for hard restart, we compare three conditions:
(i) \textbf{Mono}: same-language reboot suffix (English, as in main experiments),
(ii) \textbf{Switch}: multilingual reboot suffix (Chinese for English prompts),
(iii) \textbf{High-T}: higher-temperature resampling without suffix.

\paragraph{Preliminary findings.}
Code-switching provides marginal improvements over same-language restarts on some model--task combinations, but the effect is inconsistent across scales. In TableTable~\ref{tab:codeswitch}, we observe that the primary driver of recovery is the truncation point selection (cutting before the trap), with the choice of restart operator contributing a secondary effect. Given this, our main experiments use same-language suffixes for simplicity and reproducibility.

\section{Correlation between Escape Probability and Correctness in Trap Cases}
\label{app:escape_dist}
\begin{figure}[h]
    \centering
    \includegraphics[width=0.5\columnwidth]{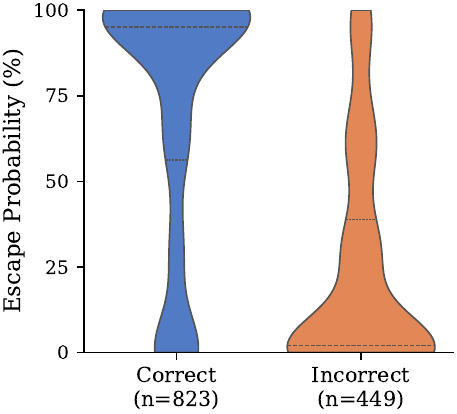}
    \caption{Correlation between Escape Probability and Correctness in Trap Cases}
    \label{fig:escape_violin}
\end{figure}

The predicted escape probability $\hat{p}$ is discriminative (Figure~\ref{fig:escape_violin}): correct trajectories concentrate at high $\hat{p}$ , while incorrect ones spread toward lower values.
